\documentclass{article}
\pdfoutput=1
\usepackage{mnscrpt_arxiv}
\usepackage{graphicx}
\usepackage[utf8]{inputenc} 
\usepackage[T1]{fontenc}    
\usepackage{hyperref}       
\usepackage{url}            
\usepackage{booktabs}       
\usepackage{amsfonts}       
\usepackage{nicefrac}       
\usepackage{microtype}      
\usepackage{multirow}       
\usepackage[round]{natbib}
\usepackage{makecell}

\newcommand{\orcid}[1]{\href{https://orcid.org/#1}{\includegraphics[width=8pt]{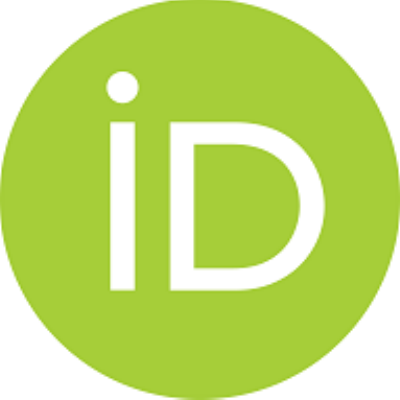}}}

\title{Modeling differential rates of aging using routine laboratory data; Implications for  morbidity and health care expenditure}

\author{
  Alix Jean~Santos \orcid{0000-0001-7490-3761}\\
  OptumLabs \\
  Minnetonka, MN\\
  \texttt{alixsantos@uhg.com} \\
   \And
 Xavier Eugenio~Asuncion \orcid{0000-0002-7362-2260}\\
  OptumLabs\\
  Minnetonka, MN\\
  \texttt{xavierasuncion@uhg.com} \\
   \AND
 Camille~Rivero-Co \orcid{0000-0001-6162-3859} \\
  OptumLabs \\
  Minnetonka, MN \\
  \texttt{camillerivero@uhg.com} \\
   \And
   Maria Eloisa~Ventura \orcid{0000-0002-9548-548X} \\
   OptumLabs \\
   Minnetonka, MN \\
   \texttt{mariaventura@uhg.com} \\
   \And
   Reynaldo~Geronia II \orcid{0000-0002-1054-3880}\\
   OptumLabs \\
   Minnetonka, MN \\
   \texttt{reynaldogeronia@uhg.com} \\
   \And
   Lauren~Bangerter \orcid{0000-0001-8452-2495}\\
   OptumLabs \\
   Minnetonka, MN \\
   \texttt{laurenbangerter@uhg.com} \\
   \And
   Natalie~E.~Sheils\ \orcid{0000-0002-0631-5834} \thanks{Corresponding author} \\
   OptumLabs \\
   Minnetonka, MN  \\
   \texttt{nataliesheils@uhg.com} \\
}

\begin{document}
\maketitle

\begin{abstract}
Aging is a multidimensional process where phenotypes change at varying rates. Longitudinal studies of aging typically involve following a cohort of individuals over the course of several years. This design is hindered by cost, attrition, and subsequently small sample size. Alternative methodologies are therefore warranted. In this study, we used a variational autoencoder to estimate rates of aging from cross-sectional data from routine laboratory tests of 1.4 million individuals collected from 2016 to 2019. By incorporating metrics that would ensure model's stability and distinctness of the dimensions, we uncovered four aging dimensions that represent the following bodily functions: 1) kidney, 2) thyroid, 3) white blood cells, and 4) liver and heart. We then examined the relationship between rates of aging on morbidity and health care expenditure. In general, faster agers along these dimensions are more likely to develop chronic diseases that are related to these bodily functions. They also had higher health care expenditures compared to the slower agers. K-means clustering of individuals based on rate of aging revealed that clusters with higher odds of developing morbidity had the highest cost across all types of health care services. Results suggest that cross-sectional laboratory data can be leveraged as an alternative methodology to understand age along the different dimensions. Moreover, rates of aging are differentially related to future costs, which can aid in the development of interventions to delay disease progression.
\end{abstract}


\section{Introduction}
Biological aging is a process that occurs when the integrity of biological systems starts to decline gradually and progressively due to the accumulation of unrepaired cellular and molecular damage \citep{Lopez-Otin-2013, Kirkwood-2005}. This decline in integrity is associated with structural and functional changes in an organ that translate to increased risk for various chronic conditions \citep{Kennedy-2014, Denic-2016, Kim-2015}. Thus, finding a metric that characterizes biological aging that is robust, precise, reliable, and sensitive to change is key to understanding the aging process and the mechanisms that explain the coexistence of age-related chronic diseases \citep{Ferrucci-2019}.

Various computational methods have also been developed to estimate the biological age of humans from biomarkers that measure physical, physiological, and biochemical parameters of the human body. For example, \citeauthor{W-2014} used principal component analysis (PCA) to create a formula for biological age using seven biomarkers that reflect the function of the vital organs of 505 Chinese participants with age range of 35 to 91 years \citep{W-2014}. They validated their methodology using telomere lengths and chronological age as proxy measures for biological age. \citeauthor{Yoo-2017} also implemented PCA on 15 biomarkers taken from 557,940 Koreans aged 20--93 years old \citep{Yoo-2017}. These biomarkers, surveyed from 1994 to 2011, included measurements commonly ordered in a general health check-up such as systolic blood pressure, blood urea nitrogen (BUN), and fasting blood sugar. Their study showed that biological age could be used as a promising predictive marker for mortality based on their Cox proportional-hazards model without requiring invasive and expensive procedures. Similarly, Belsky and colleagues developed a method to quantify an individual's rate of aging from a single blood test by training an elastic net regression model on rates of change in 18 biomarkers across 12 years to predict methylation of DNA \citep{belsky_2020}. The model was created from 954 New Zealanders (aged 26--38 years) who participated in the Dunedin Study. They found that individuals with faster rates of aging had greater risk of developing chronic diseases and early death, while those who aged slower performed better on physical and mental tasks. While these studies show promising approaches to quantifying biological age, this work is generally limited by using a specific cohort of individuals or having a small sample size. Furthermore, most of the previously mentioned studies modeled aging under the assumption that its trajectory is along a single dimension. However, the differences on how individuals evolve as they age suggest aging does not progress on a single-dimensional trajectory \citep{karol_2009, carmona-2016}.

\citeauthor{pierson2019inferring} proposed a model that describes how individuals' phenotypes evolve over time by using low-dimensional latent state vector representation with components that correspond to the aging rates along multiple dimensions \citep{pierson2019inferring}. Without using large-scale longitudinal data, the authors designed a variational autoencoder (VAE) that represents the phenotypes as a function of a linearly evolving state that allows for modeling of the rates of aging by only using single-time-point measurements (i.e., cross-sectional data) of the phenotypes across different individuals. Their VAE model learned multidimensional rates of aging using cross-sectional laboratory, anthropometric, and physiological measurements, and uncovered an interpretable, low-dimensional representation of how human phenotypes change with age. Pierson and colleagues linked these dimensions with mortality, aging risk factors, and conditions (e.g., hypertension, rheumatoid arthritis, and diabetes). 

The generalizability of the methodology presented by \citeauthor{pierson2019inferring} however, has not been validated. The model was developed and evaluated on a rather limited population, using data from 270,000 volunteers within a limited age range of 40--69 from the UK Biobank \citep{Sudlow-2015}. The relatively curated data from the structured biobank study were collected from about 30 testing centers throughout the United Kingdom, with most biomarkers per individual measured on the same day which is often not the case in practice. Moreover, applying the model to a cohort under a different healthcare system such as the United States is an important step in validating their approach. 

This study addressed the limitations of the methodology of \citeauthor{pierson2019inferring} while testing the scalability of the model on a significantly larger and more unrefined dataset. Our cohort is composed of 1.4 million individuals, around five times the population used in the referenced study, with an age range of 40--106 years old. The model's robustness against inherent differences in data generation and reporting was also effectively tested since the raw data was produced from the normal operation of testing centers across the US, which varies from routine physical exams to the diagnosis of mild or severe ailments. We built cross-sectional data of individuals in the cohort to span one month--a more realistic and less restrictive period than a single day. Most importantly, we introduced changes to improve the ease of adoption and value of the model. We investigated a modified set of phenotypes limited to those captured by laboratory tests commonly availed during general health check-ups. We introduced additional model selection metrics and steps that maximize the insights to be gained from the aging dimensions, ensure stability of results, and allow flexibility for incorporating inputs from domain experts in optimizing the aging model. We also examined the relationships of the different aging dimensions to health care expenditure in addition to its association with various chronic diseases.

\section{Methodology}
\label{sec:methodology}
\subsection{Data processing}
We used de-identified laboratory and claims data of Medicare and commercially insured enrollees in a research database from a single large US health insurance provider (the UnitedHealth Group Clinical Discovery Database). The subpopulation of interest includes individuals 40 years old and older with numeric laboratory test results within January 2016 to June 2020. We excluded individuals with pregnancy-related claims to avoid using atypical laboratory results during pregnancy in deriving the aging rates. These criteria resulted to a preliminary cohort of 12 million people, 1.4 billion unique laboratory records, and 7,728 unique laboratory-tested human physical components. Each enrollee had only a few components tested, with less than 10$\%$ having results on the most tested component. Thus, sensitivity analyses were done to optimize the trade-off between using more components as model features and having a substantially larger cohort. Considering the most tested components and aggregating them monthly, we derived a dataset composed of 1,389,280 individuals  for building and optimizing the aging model. The dataset was equivalent to 65,296,160 laboratory test results and spanned 47 laboratory components that were all tested within a month per person in the cohort. All laboratory components were assayed from blood tests which include the complete blood count panel and tests for  thyroid, lipid, metabolic, kidney and liver function. The final cohort is composed of 55.5$\%$ females covering an age range of 40 to 106 years old. The description of the 47 laboratory-tested components used as model features in inferring the aging rates and the details on how the de-identified laboratory data were cleaned are provided as supplementary information. We used the earliest available cross-sectional data of an individual so that we could evaluate the model based on the diagnoses made after the laboratory tests were taken.

De-identified claims data were used to interpret the derived aging rate dimensions by examining their relationships with morbidity and health care spending. The database contains medical (doctor, inpatient, outpatient) and pharmacy claims for services submitted for third party reimbursement. Medical and pharmacy claims are available as International Classification of Diseases, Tenth Revision, Clinical Modification (ICD-10-CM) and National Drug Codes (NDC) claims, respectively. Expenditure analyses were restricted to costs incurred by individuals who were continuously enrolled for 12 months and had complete cost data for four years following the month of the laboratory test results. Finally, commercial enrollees who were not fully insured were dropped from the expenditure analyses to improve the accuracy of our cost calculations.

\subsection{Model training and selection}
\subsubsection{Aging model and hyperparameter tuning}

The VAE-based aging model proposed by \citeauthor{pierson2019inferring} assumes that each person has phenotypes that evolve over time across multiple dimensions. In our work, these phenotypes correspond to the 47 laboratory-tested human physical components. The aging model represents the components of an individual as a function of a linearly evolving latent state. The values of the components at time $t$, feature vector $\mathbf{x}_t$, are modeled as a function of $t$, rate vector $\mathbf{x}$, and a normally distributed bias term $\epsilon^\prime$ that encodes the non-time-varying information:

\begin{equation}
    \mathbf{x}_t = f(\mathbf{r}t)+\epsilon^\prime
\end{equation}

\noindent In the above equation, $\mathbf{r}$ is a multidimensional vector that contains the information on how fast an individual's features progress with $t$ along different dimensions. Consequently, the value $\mathbf{r}t$ represents the biological age of an individual, per dimension, at a certain point in time. Thus, by using data from a large group of individuals with varying ages, the model learns how to extract these biological ages without using longitudinal information. Specifically, the goal of the VAE model is to estimate, using available cross-sectional data $\mathbf{X}$ and age vector $t$, the parameters of $f$ and $\epsilon^\prime$, and in turn extract the rate matrix $\mathbf{R}$. This matrix describes the rate information and contains a unique vector $\mathbf{r}$ for each individual, which is assumed to not change with time. 

To generate $\mathbf{R}$ using the above process, the following hyperparameters are incorporated in the VAE model: the size of the rate vector $n$, the aging rate scaling factor $\sigma_r$, the number of monotone features, and the polynomial fitting degrees $S$. The size of $\mathbf{r}$ defines the number of aging dimensions to be extracted. The factor $\sigma_r$ scales the spread of the log of the aging rates closer to a biologically realistic distribution, with $\mathbf{r}$ assumed to have a prior log-normal distribution. The number of monotone features sets which components, based on their Spearman's rank correlations with age, are defined to be monotonic with time. Specifying the monotone features is necessary to make the function $f$, and hence the rates of aging $\mathbf{r}$, identifiable. If a component is monotone, a series of transformations is performed when processing the latent features back to their original component values. These include a linear transformation parametrized by fitting a polynomial function with degrees $S$. If a component is non-monotone, the decoder portion of the network is used to reconstruct the components from the latent features. Other hyperparameters to control the deep learning process are also used in training the VAE model. The reader is referred to the paper of \citeauthor{pierson2019inferring} for more detailed information \citep{pierson2019inferring}. 

We built the aging model by using standard-scaled data from 80$\%$ of the individuals as inputs for training and the remaining 20$\%$ for optimization via hyperparameter tuning. In this study, we explored two sets of combinations of hyperparameter values wherein the second set had a wider range of values compared to the first (Table \ref{table:hyperparameters}). We set $\sigma_r$ to $0.1$ and $S$ to $[1/5, 1/4, 1/3, 1/2, 1, 2, 3, 4, 5]$ as in the model by \citeauthor{pierson2019inferring} For model training, we used a batch size of $4,096$, maximum epochs of $300$, ReLU activation function, and an Adam optimizer.

\begin{table}[t]
 \caption{Hyperparameter values of the tuning search space and of the model by
 \citeauthor{pierson2019inferring}}
 \label{table:hyperparameters}
\begin{tabular}{lccc}
\toprule
Hyperparameter & Set 1 & Set 2 & \citeauthor{pierson2019inferring} \\
\toprule
Size of rate vector & $3, 4$ & $3, 4, 5, 6$ & $5$ \\
Number of monotone features$^*$ & $34, 28, 18, 9, 3$ & $34, 28, 18, 9, 3$ & $45$ \\
Network architecture$^{**}$ & 
    \makecell{$\left\{36, 34, 32, 30, 28\right\}-$ \\ $\left\{24, 22, 20, 18, 16\right\}$} & 
    \makecell{$\left\{64\right\} -\left\{32, 16\right\}$, \\
        $\left\{128\right\} - \left\{32\right\}$, \\ 
        $\left\{32\right\}-\left\{16\right\}$} & 
    $\left\{50\right\} - \left\{20\right\}$ \\
Learning rate & $0.0005, 0.0006, 0.0007$ & \makecell{$0.0001, 0.0003, 0.0005,$ \\ $ 0.0007, 0.0010$} & $0.0005$ \\
\bottomrule
\multicolumn{4}{l}{\footnotesize $^*$Based on the number of features whose Spearman's rank correlation with age have magnitudes $\geq$ 0.05, 0.10, 0.15, 0.20, and 0.25} \\
\multicolumn{4}{l}{\footnotesize $^{**}$\{sizes of outer layer\} -- \{sizes of inner layer\} of encoder and decoder}   
\end{tabular}
\end{table}

\subsubsection{Selection criteria}
We employed both quantitative and qualitative methods to select the best VAE aging model. The quantitative metrics we used for optimizing the hyperparameters are the following: (1) reconstruction correlation, (2) intra-model similarity, and (3) inter-model similarity. Reconstruction correlation quantifies a model's accuracy and representational power, that is, the similarity between the original laboratory data and their corresponding decoded values from the encoded aging rates. Thus, a high value implies that a model has learned to represent the laboratory data as lower dimensional aging rates without significant loss of information. The reconstruction correlation of each trained model is measured by taking the Pearson correlation between the optimization data from the remaining 20$\%$ of the cohort and their corresponding reconstruction from the learned aging rates. 

However, it is not enough for a good model to have high accuracy. For a model to be valuable, it must also maximize the number of unique aspects of aging that it captures from the input data. In this study, we introduce the intra-model similarity metric, which measures how similar the dimensions in a single model are. A low value is desirable since it implies that the aging dimensions learned by a model are generally distinct from one another. This metric is determined as follows. First, we calculate a model's correlation matrix $\mathbf{C} = [\mathbf{c}_1\  \mathbf{c}_2\  \ldots \mathbf{c}_n]$ using the laboratory data and the aging rates $\mathbf{R} = [\mathbf{r}_1\  \mathbf{r}_2\  \ldots \mathbf{r}_n]$ it learned from the data. Matrix $\mathbf{C}$ describes the correlation of rate vectors $r_j$ in $\mathbf{R}$ with the $m$ components used as features in training the model. From $\mathbf{C}$, we then identify which features in each dimension satisfy a predefined threshold $\delta$ such that $|\mathbf{c}_j|\geq\delta$, where $0\leq\delta\leq 1$. These sets of features, represented by a binary matrix $\mathbf{B}$, are then used to calculate the similarity between all possible pairs of rate dimensions. This value is calculated pairwise by dividing the number of features satisfying the threshold in both dimensions by the number of features satisfying the threshold in either dimension. Taking the average of pairwise similarity scores relevant to each dimension gives us a vector of similarity scores $s$, which is finally averaged to obtain the model's intra-model similarity. In this study, the predefined threshold used was $\pm 0.35$. 

\begin{figure}[ht]
	\centering
		\includegraphics[width=0.75\textwidth]{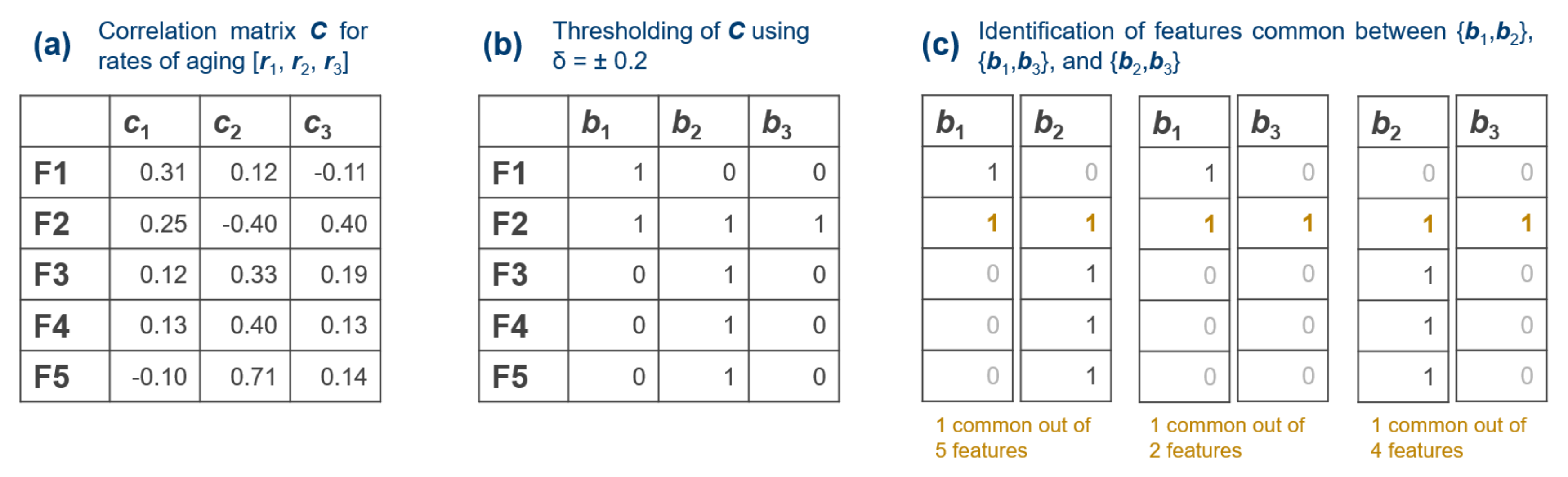}
	\caption{Diagram showing the calculation of intra-model similarity from a sample correlation matrix with $m = 5$ features and $n = 3$ rate dimensions.}
	\label{FIG:intramodelsim}
\end{figure}

We illustrate this procedure in Figure \ref{FIG:intramodelsim}. Step (a) shows a dummy correlation matrix from a model with five features and three rate dimensions. Applying a threshold of $\delta = 0.2$ to these values results to a binary matrix in (b). Dimensions are then compared pairwise in (c), revealing that only the second feature had a correlation that satisfied the threshold in all three dimensions. On the other hand, each pair of dimensions had different total counts of features that met the threshold in either dimension: five for $\mathbf{b}_1$ and $\mathbf{b}_2$, two for $\mathbf{b}_1$ and $\mathbf{b}_3$, and four for $\mathbf{b}_2$ and $\mathbf{b}_3$. This results in pairwise similarity scores for $\left\{\mathbf{b}_1, \mathbf{b}_2\right\}$, $\left\{\mathbf{b}_1, \mathbf{b}_3\right\}$, and $\left\{\mathbf{b}_2, \mathbf{b}_3\right\}$ of $0.20$, $0.50$, and $0.25$, respectively. The similarity score of each dimension is then calculated by a simple average of relevant pairwise scores: for example, the similarity score for $\mathbf{b}_1$ is the average of pairwise scores for $\left\{\mathbf{b}_1, \mathbf{b}_2\right\}$ and $\left\{\mathbf{b}_1, \mathbf{b}_3\right\}$, or $0.350$. This leads to $s = [0.350, 0.225, 0.375]$ and hence an intra-model similarity of $0.475$.

Lastly, a model derived from a non-deterministic deep learning algorithm must generate consistent insights that are independent from the algorithm's initialization state for it to be considered robust. Thus, we introduce the inter-model similarity metric to quantify model stability. The metric describes how models, using the same set of hyperparameters but trained using different machine learning seeds, are alike with respect to learned correlations. Calculating a model's inter-model similarity resembles the process for intra-model similarity, the only difference being that the dimensions being compared pairwise come from different correlation tables generated using ten different initialization seeds. It is more desirable for this metric to have a high value as it denotes greater stability.

\subsubsection{Selection process}
There were 346 models successfully built from the hyperparameter search. Even models with very high reconstruction accuracies manifested a wide range of intra-model similarities, further justifying the need to control the latter metric. There were $206$ models that satisfied our preset reconstruction criterion of at least $0.85$. Among these, we chose three models for each of $\{3, 4, 5, 6\}$ aging dimensions that had the lowest intra-model similarity. The resulting 12 models were then evaluated qualitatively to incorporate informed choices from domain experts and literature. This step allows us to narrow down our models to those that produced the most interpretable results based on the characterization of the dimensions and clustering analysis. (These are described in the succeeding sections in more detail.) Table \ref{table: top models} shows the resulting metrics of the most interpretable model/s per number of aging dimensions. We note that lower learning rate values ($\leq 0.0005$) and wider network architectures (64 encoder nodes and/or 32 decoder nodes) generally resulted to models with better metrics. The size of rate vector and the number of monotone features both did not affect the metrics of the generated models in a consistent manner, but the top candidate models all had 34 monotone features, which was the highest number of monotone features tested for model tuning.

\begin{table}[ht]
    \caption{Resulting metrics of top candidate models.}
    \label{table: top models}
    \begin{tabular}{cccc|ccc}
        \toprule
        \shortstack{Size of \\ rate vector} & 
        \shortstack{Monotone \\ features} & 
        \shortstack{Network \\ architecture} &
        \shortstack{Learning  \\ rate} & 
        \shortstack{Reconstruction  \\  correlation} & 
        \shortstack{Intra-model \\ similarity (\%)} & 
        \shortstack{Inter-model \\ similarity (\%)} \\
        \toprule
        3 & 34 & {[}64 - 32{]} & 0.0003 & 0.963 & 0.00 & 93.9 \\
        \textbf{4} & \textbf{34} & \textbf{{[}64 - 16{]}} & \textbf{0.0001} & 
        \textbf{0.868} & \textbf{5.56} & \textbf{92.2} \\
        5 & 34 & {[}64 - 32{]} & 0.0003 & 0.972 & 0.00 & 93.3 \\
        5 & 34 & {[}64 - 32{]} & 0.0001 & 0.960 & 0.00 & 94.4 \\
        6 & 34 & {[}64 - 16{]} & 0.0003 & 0.889 & 3.00 & 88.6 \\ 
        \bottomrule
        \multicolumn{7}{l}{\footnotesize Note: Selected optimal model based on both quantitative and qualitative metrics is presented in boldface.}  
    \end{tabular}
\end{table}

\begin{figure}[ht]
	\centering
		\includegraphics[width=0.8\textwidth]{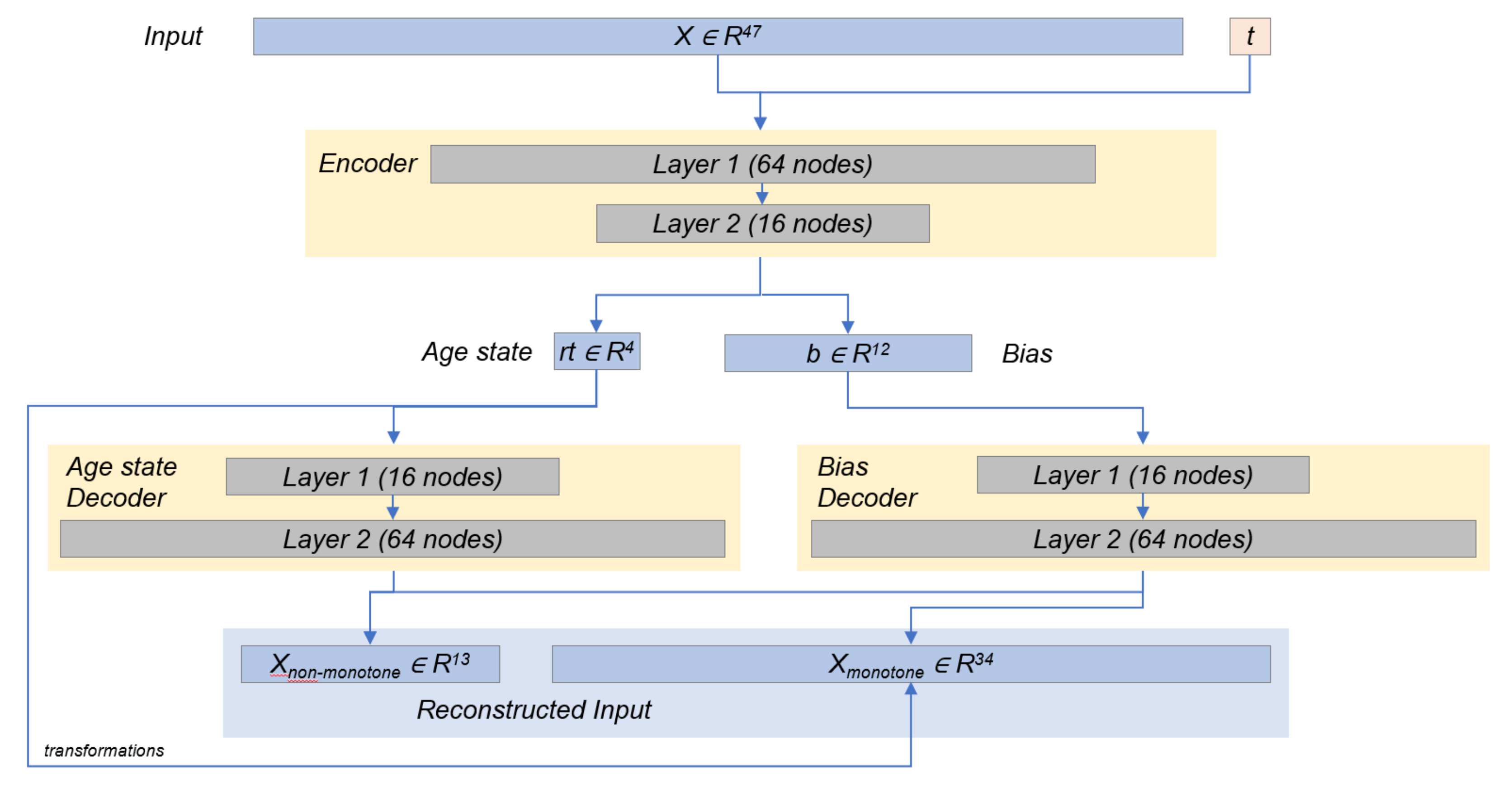}
	\caption{Diagram of VAE aging model with optimized hyperparameters}
	\label{FIG:vae}
\end{figure}

After further qualitative evaluation, we selected the model with 4 aging dimensions, 34 monotone features, 64 nodes in the decoder layer, 16 nodes in the encoder layer, and 0.0001 learning rate as the optimal aging model. This model had a reconstruction correlation of 0.868, an intra-model similarity of 5.56\%, and an inter-model similarity of 92.2\%. The final model architecture is illustrated in Figure \ref{FIG:vae}. 

\subsection{Characterization of aging rate dimensions}
The aging rate dimensions produced by the model were then characterized based on their correlations to the components and associations to historical and future diagnoses. Components that most closely correspond to each aging dimension were identified by comparing the Pearson correlation values between dimension-component pairs. Meanwhile, associations between aging rate dimensions and historical diagnoses were examined by building ordinary least squares (OLS) models for each dimension, using the presence of chronic diseases as independent variables and the aging dimension as dependent variable. On the other hand, chronic diseases that have high odds of eventually developing in fast agers in each dimension were determined by calculating from logistic regression models the odds ratio of a disease per aging dimension. We used the Chronic Condition Data Warehouse (CCW) to identify the chronic diseases \citep{cmsccw}. The logistic regression models used the aging dimensions as covariates and the presence of a diagnosis for the disease per individual in the succeeding years following their laboratory data as the outcome variable. In building the logistic regression models, we removed individuals who had been diagnosed with the disease prior to the month of their corresponding data. Both the OLS and logistic regression models were controlled for age and sex.

To determine how aging rates are related to future health care expenses, we compared the costs incurred by fast and slow agers along each aging dimension. We also compared fast agers in one dimension to fast agers in other dimensions. Fast and slow agers were defined as individuals who have rates of aging in the $\geq$ 90th percentile and in the $\leq$ 10th percentile of a specific dimension, respectively. Note that fast and slow agers are specific to each aging dimension; a fast ager in one dimension may not necessarily be a fast ager in another dimension. Statistical tests were performed on the cumulative costs per individual, incurred from one to four years after their laboratory results were used for model building. The distribution of costs incurred by slow and fast agers were compared using the Mann-Whitney rank test. Meanwhile, the distribution of costs incurred by fast agers across the aging dimensions were compared using the Kruskal-Wallis test followed by a pairwise Conover's squared rank test with Holm-Bonferroni correction.

\subsection{Clustering of individuals based on aging rates}
Aside from characterizing the aging rate dimensions themselves, we also used the rates to group individuals who exhibited similar rates of aging across all dimensions via K-means clustering. The final number of clusters was identified through elbow method, silhouette analysis, and other heuristics which involved examining whether distribution of aging rates are distinct across the clusters. The clusters were afterwards compared in terms of their association to future diagnoses and their corresponding health care costs. The associated diagnoses were determined by calculating the cluster's odds ratios for various diseases against all other clusters. In the calculation of the odds ratios, individuals who had been previously diagnosed with the disease were excluded in the analysis. The OLS and logistic regression models used were controlled for age and sex.

\section{Results}
\subsection{Dimensions of rates of aging}
\subsubsection{Correlated components and associated diseases}
\begin{figure}[ht]
	\centering
	\includegraphics[width=0.5\textwidth]{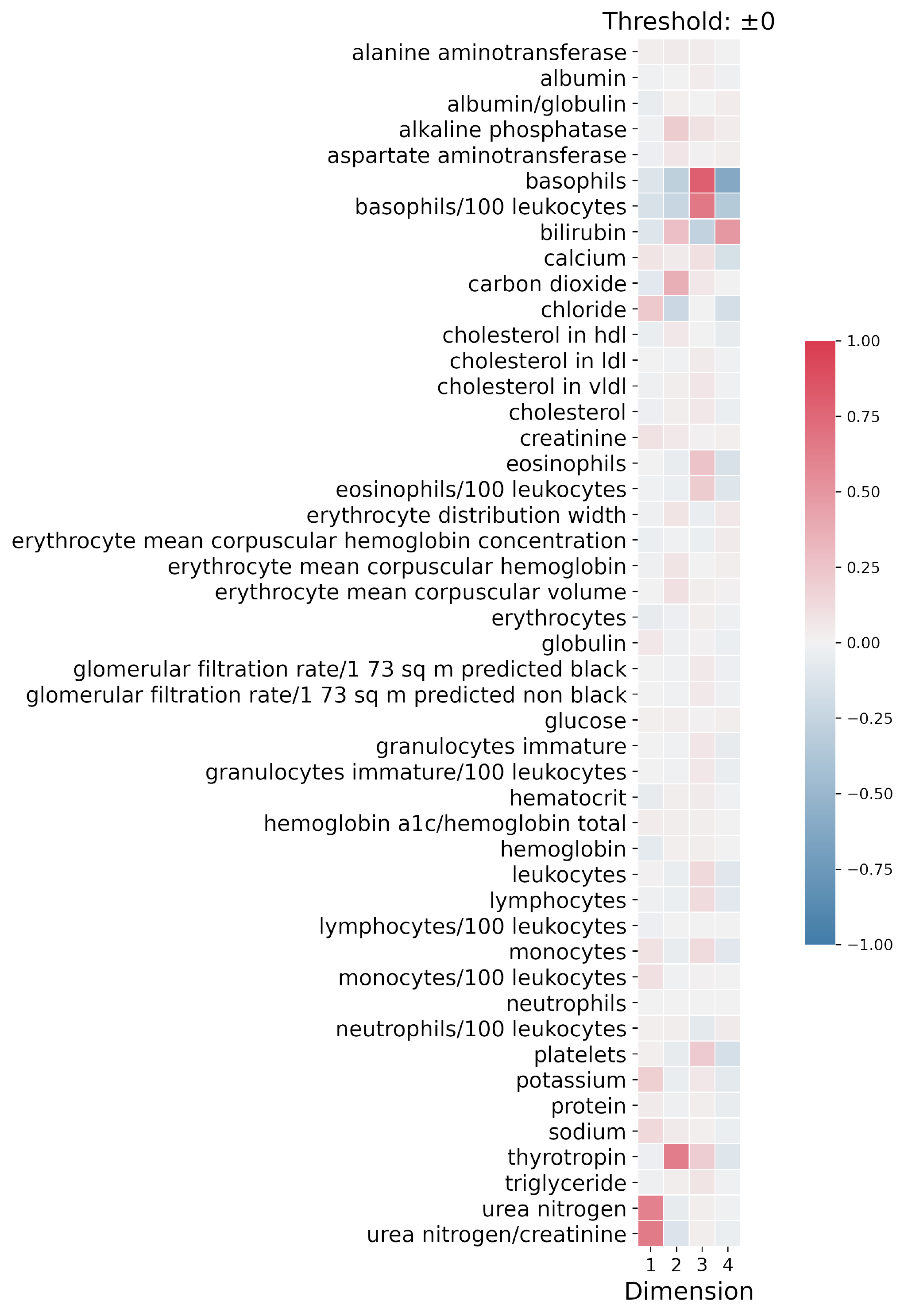}
	\caption{Pearson correlations between aging dimensions and components}
	\label{FIG:pearson_corr}
\end{figure}

The aging dimensions were distinguished from one another based on the most correlated laboratory-tested components. In our analyses, we limited the components to those whose Pearson correlations with the aging dimensions had absolute values of at least 0.35, selected arbitrarily. As shown in Table \ref{table: r_associations}, the first dimension of the aging rate matrix $(r_1)$ increases with BUN and BUN-to-creatinine ratio (BCR), the second $(r_2)$ with thyrotropin (TSH) and carbon dioxide (CO$_2$), and the third $(r_3)$ with the absolute and relative basophil count; the fourth $(r_4)$ increases with increasing bilirubin and decreasing absolute basophil count. The correlations of the dimensions with other features are presented as supplementary information (Figure \ref{FIG:pearson_corr}). It is noted that $r_1$ was also weakly correlated to creatinine and kidney-regulated electrolytes (sodium, chloride, potassium, and calcium), while $r_3$ was weakly correlated to total leukocytes and its subtypes (except basophils). Also presented in Table \ref{table: r_associations} are the historical chronic diagnoses associated to each aging dimension as uncovered by the OLS models.

\begin{table}[ht]
\caption{Strongly correlated components and positively associated historical diagnoses to aging dimensions.}
\label{table: r_associations}
\begin{tabular}{@{}clcl@{}}
\toprule
Aging Dimension & Correlated laboratory-tested components$^\dagger$ & Correlation coefficient & Associated diagnoses \\
\toprule
    $r_1$ &
    \begin{tabular}[c]{@{}l@{}}
        Urea nitrogen \\ Urea nitrogen/creatinine
    \end{tabular} &
    \begin{tabular}[c]{@{}l@{}}
        +0.611 \\ +0.646
    \end{tabular} &
    \begin{tabular}[c]{@{}l@{}}
        End-stage renal disease* \\ Chronic kidney disease* \\ Endometrial cancer* \\ Myocardial infarction* \\ Obstructive sleep apnea*
    \end{tabular} \\
\midrule
    $r_2$ &
    \begin{tabular}[c]{@{}l@{}}
        Thyrotropin \\ Carbon dioxide
    \end{tabular} &
    \begin{tabular}[c]{@{}l@{}}
        +0.626\\ +0.360
    \end{tabular} &
    \begin{tabular}[c]{@{}l@{}}
        End-stage renal disease* \\ Alcohol abuse* \\ Hypothyroidism* \\ Heart failure* \\ Endometrial cancer*
    \end{tabular} \\
\midrule
    $r_3$ &
    \begin{tabular}[c]{@{}l@{}}
        Basophils \\ Basophils/100 leukocytes
    \end{tabular} &
    \begin{tabular}[c]{@{}l@{}}
        +0.797\\ +0.660
    \end{tabular} &
    \begin{tabular}[c]{@{}l@{}}
        End-stage renal disease* \\ Asthma* \\ Hypothyroidism* \\ Myocardial infarction* \\ Anxiety*
    \end{tabular} \\
\midrule
    $r_4$ &
    \begin{tabular}[c]{@{}l@{}}
        Bilirubin \\ Basophils
    \end{tabular} &
    \begin{tabular}[c]{@{}l@{}}
        +0.484\\ -0.615
    \end{tabular} &
    \begin{tabular}[c]{@{}l@{}}
        Atrial fibrillation\\ Alcohol abuse \\ Heart failure \\ Alzheimer's and dementia \\ Colorectal cancer
    \end{tabular} \\
\bottomrule
\multicolumn{4}{l}{
    \begin{tabular}[c]{@{}l@{}}
        \footnotesize$^\dagger$Laboratory-tested components whose Pearson's correlation with the aging dimensions have absolute values of at least 0.35. \\
        \footnotesize$^{**}$Diseases with asterisks are associated with $\geq 1\%$ change in rate of aging.
    \end{tabular}
}                                                     
\end{tabular}
\end{table}

We also investigated whether fast agers in each dimension have higher odds of developing the associated diseases in the future. Table \ref{table:future_diag} shows the diseases that were prevalently diagnosed among fast agers within a few years after their laboratory measurement. These associations were mostly consistent with what we had found for historical diagnoses (Table \ref{table: r_associations}).

\begin{table}[ht]
\caption{Future diagnoses that are prevalent among fast agers on each rate dimension.}
\label{table:future_diag}
\begin{tabular}{@{}cllc@{}}
\toprule
Aging dimension & Correlated laboratory-tested components$^{\dagger}$ & Associated diagnoses$^{\ddag}$ & Odds ratio$^{\dagger\dagger}$ \\ \toprule
    $r_1$ &
    \begin{tabular}[c]{@{}l@{}}
        Urea nitrogen \\ Urea nitrogen/creatinine
    \end{tabular} &
    \begin{tabular}[c]{@{}l@{}}
        End-stage renal disease \\ Anemia \\ Chronic kidney disease \\
        Heart failure \\ Myocardial infarction \\ Diabetes \\ Ischemic heart disease \\ Obstructive sleep apnea \\ Hypertension \\ Arthritis
    \end{tabular} &
    \begin{tabular}[c]{@{}l@{}}
        7.03 \\ 2.45 \\ 2.22 \\
        2.12 \\ 1.87 \\ 1.72 \\ 1.51 \\ 1.51 \\ 1.44 \\ 1.33
    \end{tabular} \\
\midrule
    $r_2$ & 
    \begin{tabular}[c]{@{}l@{}}
        Thyrotropin \\ Carbon dioxide
    \end{tabular} &
    \begin{tabular}[c]{@{}l@{}}
        Hypothyroidism \\ Alcohol abuse \\ Heart failure \\
        Obstructive sleep apnea \\ Lung cancer \\ Anemia \\ 
        Chronic kidney disease \\ Myocardial infarction \\ 
        Asthma \\ Chronic obstructive pulmonary disease \\ Cataract \\
        Ischemic heart disease
    \end{tabular} &
    \begin{tabular}[c]{@{}l@{}}
        3.00 \\ 2.34  \\ 1.61 \\ 1.57 \\ 1.45 \\ 1.38 \\ 1.34 \\
        1.28 \\ 1.19 \\ 1.17 \\ 1.14 \\ 1.11 
    \end{tabular} \\
\midrule
    $r_3$ & 
    \begin{tabular}[c]{@{}l@{}}
        Basophils \\ Basophils/100 leukocytes
    \end{tabular} &
    \begin{tabular}[c]{@{}l@{}}
        Hypothyroidism \\ Chronic kidney disease \\ Heart failure \\
        Chronic obstructive pulmonary disease \\ Hypertension \\
        Anemia \\ Diabetes \\ Myocardial infarction \\ Depression \\ 
        Ischemic heart disease \\ Anxiety
    \end{tabular} &
    \begin{tabular}[c]{@{}l@{}}
        2.33 \\ 2.04  \\ 1.61 \\ 1.60 \\ 1.56 \\ 1.43 \\ 1.43 \\
        1.40 \\ 1.38 \\ 1.35 \\ 1.22
    \end{tabular} \\
\midrule
    $r_4$ & 
    \begin{tabular}[c]{@{}l@{}}
        Bilirubin \\ Basophils
    \end{tabular} &
    \begin{tabular}[c]{@{}l@{}}
        Heart failure \\ Obstructive sleep apnea \\ Hypertension
    \end{tabular} &
    \begin{tabular}[c]{@{}l@{}}
        1.26 \\ 1.23  \\ 1.21
    \end{tabular} \\
\bottomrule
\multicolumn{4}{l}{
    \begin{tabular}[c]{@{}l@{}}
        \footnotesize$^\dagger$Laboratory-tested components whose Pearson's correlation with the aging dimensions have absolute values of at least 0.35. \\
        \footnotesize$^\ddag$Only diagnoses with consistently at least 1.05 odds ratios each year from first to fourth year are shown. \\
        \footnotesize$^{\dagger\dagger}$Odds ratio calculated from data across four years.
    \end{tabular}
}
\end{tabular}
\end{table}

\subsubsection{Health care expenditure of fast agers}
About 34$\%$ of the cohort are fast agers in at least one aging dimension, while about 32$\%$ are slow agers. A subset of these people having complete medical claims data was considered for the analysis of future total services (including doctor, outpatient, and inpatient) costs. Similarly, a subset having complete pharmacy claims data was considered for the analysis of future medication costs. Moreover, extending the duration of the analysis to include more years reduced the number of individuals having complete claims data. The number of individuals considered in comparing fast to slow agers in each dimension ranged from 30 to 75 thousand per group for a one-year period and from 3.2 to 11 thousand per group for a four-year period. 

\begin{figure}[ht]
	\centering
		\includegraphics[width=0.83\textwidth]{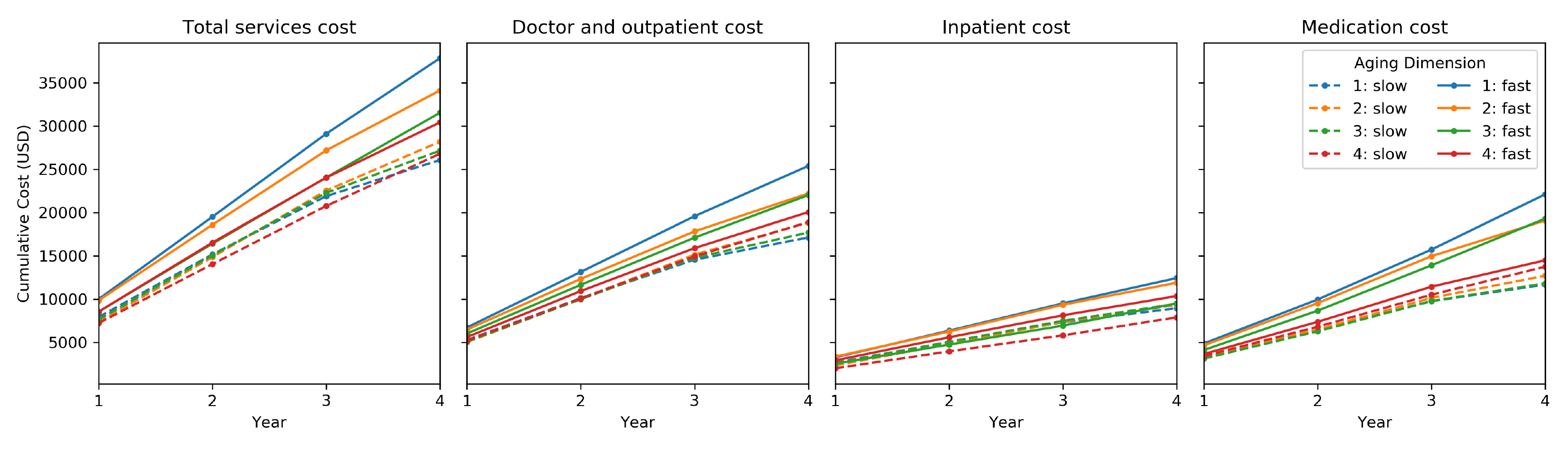}
	\caption{Mean of cumulative costs (US\$) by aging dimension and by cost type.}
	\label{FIG:cost_dimension_mean}
\end{figure}

The means of the cumulative costs incurred by fast agers are generally higher than those of slow agers in the same aging dimension (Figure \ref{FIG:cost_dimension_mean}). This is the case regardless of the cost type for all periods that we considered and in all aging dimensions, except for the inpatient costs in $r_3$ for some periods. 

\begin{figure}[ht]
	\centering
		\includegraphics[width=0.83\textwidth]{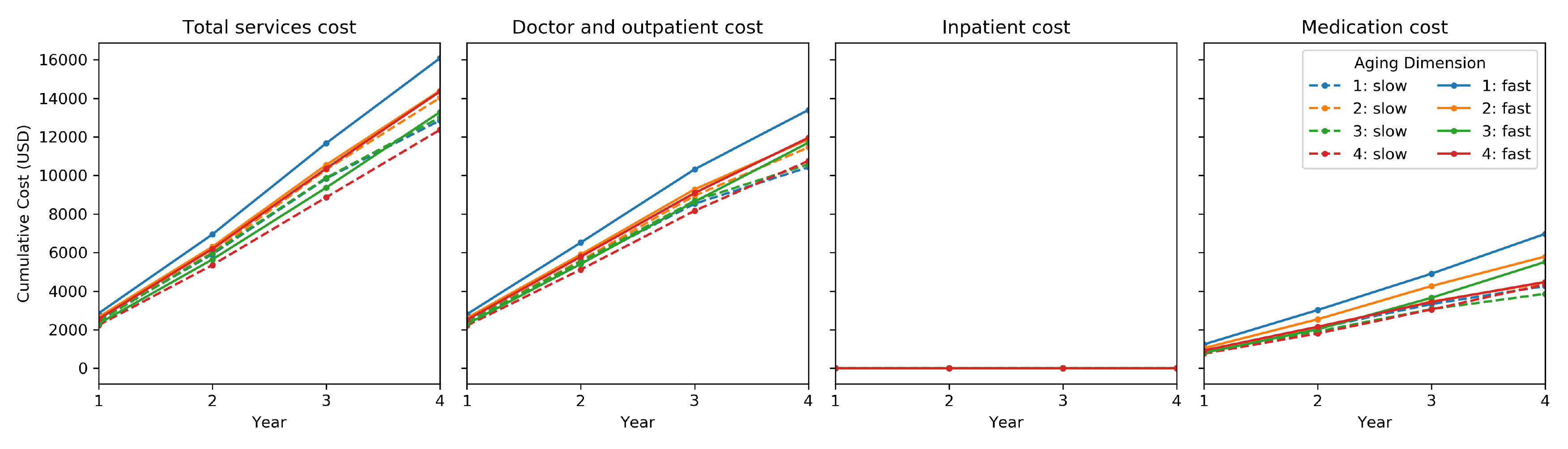}
	\caption{Median of cumulative costs (US\$) by aging dimension and by cost type.}
	\label{FIG:cost_dimension_median}
\end{figure}

We also looked at the medians and used non-parametric tests to make our analysis robust against outliers and skewed distributions (Figure \ref{FIG:cost_dimension_median}). Generally, the cumulative costs incurred by fast agers have significantly higher medians and distributions at $\alpha=0.05$ than their slow aging counterparts in the same dimension. This is the case for all aging dimensions except $r_3$, in which 64$\%$ of the slow agers are fast agers in other dimensions (55$\%$ were in $r_4$) and 67$\%$ of the fast agers are slow agers in others (50$\%$ in $r_4$). The slow agers in $r_3$ overlap mainly with fast agers of $r_4$ at a Jaccard similarity of 0.35, while fast agers in $r_3$ overlap mainly with slow agers of $r_4$ at a Jaccard similarity of 0.26. The medians of inpatient costs are mostly zero since only a few people in the cohort had to be admitted. 

\begin{figure}[ht]
	\centering
		\includegraphics[width=0.83\textwidth]{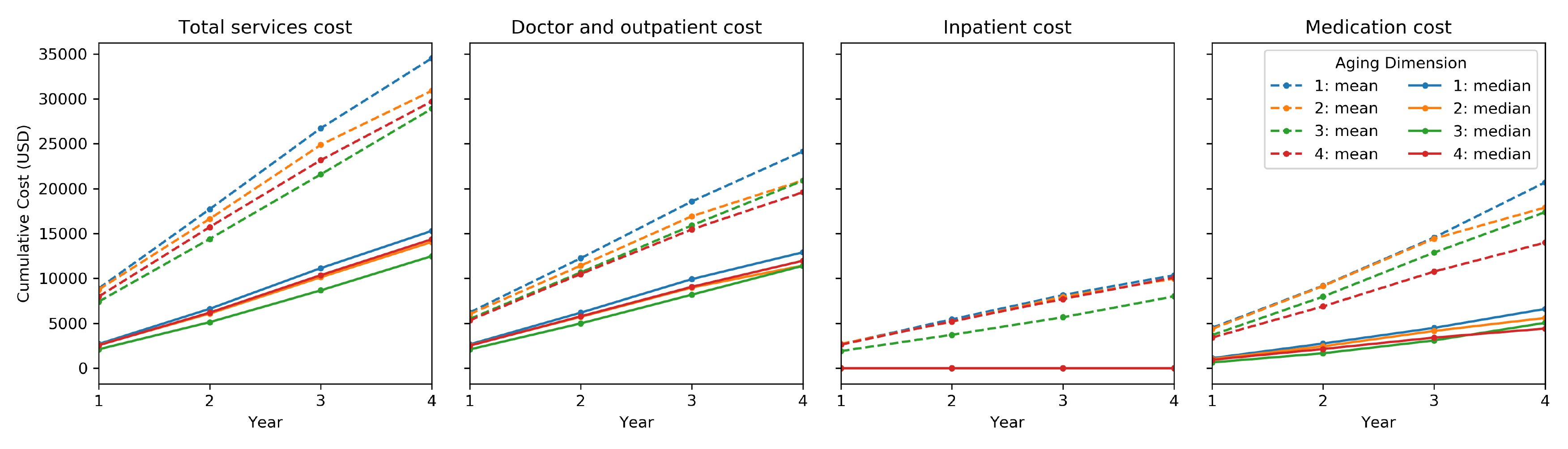}
	\caption{Mean and median of cumulative costs (US\$) of fast agers in only one aging dimension, by cost type.}
	\label{FIG:cost_dimension_fast}
\end{figure}

We also explored whether there were differences in the cumulative costs of fast agers among the four dimensions. We removed individuals who are fast agers in multiple dimensions to maintain the independence of the groups being compared in this analysis. The number of individuals ranged from about 22,000 to 49,000 per group for a one-year period and from about 2,3000 to 6,500 per group for a four-year period. The smaller subset of fast agers resulted to a downward shift in the means for all four dimensions by up to 11$\%$ (Figure \ref{FIG:cost_dimension_fast}). Similarly, the medians of the first 3 dimensions were shifted by around 5$\%$, 3$\%$, and 9$\%$ for services costs, and by about 9$\%$, 3$\%$, and 19$\%$ for medication costs, respectively. Kruskal-Wallis tests showed that the costs to be incurred by fast agers of the different dimensions were unlikely to originate from the same distribution regardless of cost type and years covered. The pairs of dimensions for which the dispersion of costs of fast agers are statistically different were determined through post-hoc Conover squared ranks test. This was then followed by Mann-Whitney tests to identify the aging dimension which were likely to have higher costs. All tests used a significance level of $\alpha=0.05$. Results showed that the distribution of costs of fast agers in each dimension (individually denoted as $C_{r_i}$) followed the general trend of  $C_{r_1} > C_{r_2} \approx C_{r_4} > C_{r_3}$ for total services category and its primary component: doctor and outpatient cost. For the inpatient cost component of total services, the trend was $C_{r_4} > C_{r_1} \approx C_{r_2} > C_{r_3}$. For the medication cost category, $C_{r_1} > C_{r_2} > C_{r_4} > C_{r_3}$. We noted that before removing individuals who are fast agers in multiple dimensions for this analysis, the fast agers in $r_2$ and $r_4$ with claims data have an unusually high Jaccard similarity of 0.12 while other pairs of dimensions only range from almost zero to 0.06. A quarter of fast agers in $r_2$ are also fast agers in $r_4$ and a fifth of fast agers in $r_4$ are also fast agers in $r_2$. 

\subsection{Clusters of individuals}
\subsubsection{Distribution of rates of aging}

\begin{figure}[ht]
	\centering
		\includegraphics[width=0.83\textwidth]{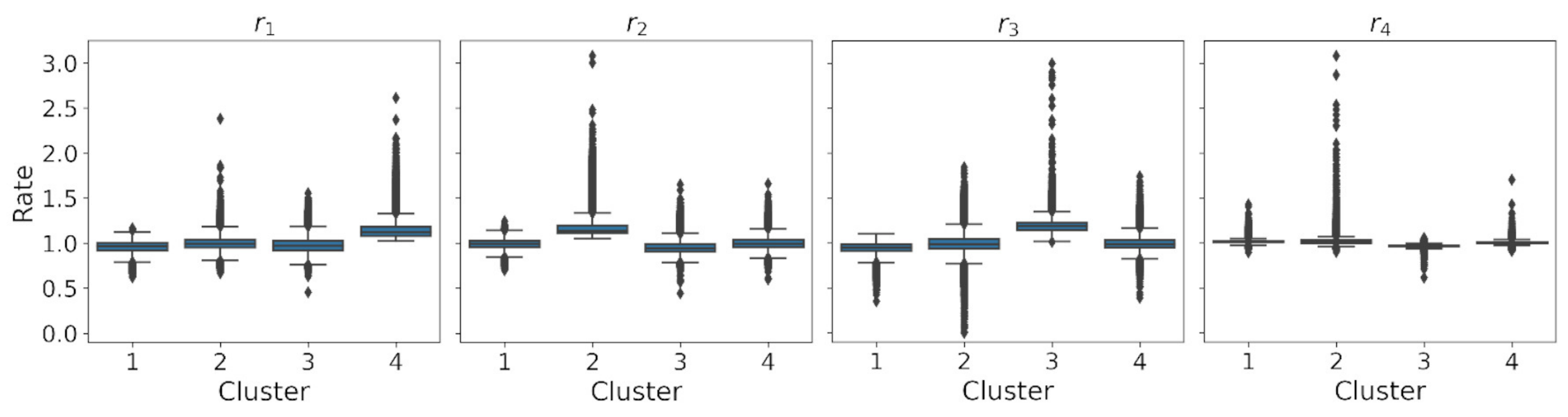}
	\caption{Distribution of clusters' rates of aging per dimension:  kidney ($r_1$), thyroid ($r_2$), white blood cells ($r_3$), and liver and heart ($r_4$).}
	\label{FIG:clusters}
\end{figure}

Grouping individuals according to their rates of aging revealed that almost all clusters can be linked back to a distinct dimension when the number of clusters was set to four. We found that except for $r_4$, there was a unique cluster consisting of members who exhibited the fastest rates of aging on each dimension (Figure \ref{FIG:clusters}). For instance, in $r_1$, cluster 4 generally aged the fastest compared to other clusters. On the other hand, cluster 2 had the fastest rates of aging along $r_2$ and very high outlier rates in $r_4$. Meanwhile, cluster 3 exhibited faster rates of aging in $r_3$, whereas it showed the slowest rates of aging in $r_4$. Of note, there was no dimension wherein the rates of aging of the individuals belonging to cluster 1 stood out.

\subsubsection{Associated future diagnoses}
We also observed that there were clusters that consistently achieved the highest odds ratios for a particular diagnosis every year (Table \ref{table:cluster_odds_future_diag}). Cluster 1 obtained the highest odds for prostate cancer. Cluster 2 had the highest odds for hypothyroidism, alcohol abuse, heart failure, atrial fibrillation, obstructive sleep apnea, Alzheimer's and dementia, hypertension, depression, and stroke. On the other hand, cluster 3 achieved the highest odds for chronic obstructive pulmonary disease (COPD), asthma, hyperlipidemia, and anxiety. Lastly, cluster 4 consistently had the highest odds for end-stage renal disease (ESRD), anemia, chronic kidney disease (CKD), endometrial cancer, myocardial infarction, ischemic heart disease, arthritis, diabetes, osteoporosis, cataract, and glaucoma.

\begin{table}[ht]
\caption{Future diagnoses that are consistently prevalent per cluster from first year to fourth year of observation.}
\begin{tabular}{@{}cllc@{}}
\toprule
    Cluster & 
    \begin{tabular}[c]{@{}l@{}}
        Related dimension$^{\dagger}$\\ (Correlated laboratory-tested components)
    \end{tabular} &
    Associated diagnoses$^\ddag$ & Odds ratio $^{\dagger\dagger}$ \\
\toprule
    1 &
      &
    \begin{tabular}[c]{@{}l@{}}
        Prostate cancer
    \end{tabular} &
    \begin{tabular}[c]{@{}l@{}}
        1.08 
    \end{tabular} \\
\midrule
    2 & 
    \begin{tabular}[c]{@{}l@{}}
        $r_2$\\ (Thyrotropin, Carbon Dioxide) \\
        $r_4$\\ (Bilirubin, Basophils)
    \end{tabular} &
    \begin{tabular}[c]{@{}l@{}}
        Hypothyroidism \\ Alcohol abuse \\ Heart failure \\
        Atrial fibrillation \\ Obstructive sleep apnea \\ Drug abuse \\
        Alzheimer's and dementia \\ Hypertension \\ Depression \\ Stroke
    \end{tabular} &
    \begin{tabular}[c]{@{}l@{}}
        2.56 \\ 1.60 \\ 1.48 \\ 1.34 \\ 1.28 \\ 1.20 \\ 1.16 \\ 1.14 \\ 1.13 \\ 1.13
    \end{tabular} \\
\midrule
    3 & 
    \begin{tabular}[c]{@{}l@{}}
        $r_3$\\ (Basophils, Basophils/100 leukocytes) \\
        $r_4$\\ (Bilirubin, Basophils)
    \end{tabular} &
    \begin{tabular}[c]{@{}l@{}}
        Chronic obstructive pulmonary disease \\ Asthma \\ Hyperlipidemia \\ Anxiety
    \end{tabular} &
    \begin{tabular}[c]{@{}l@{}}
        1.24 \\ 1.13 \\ 1.06 \\ 1.06
    \end{tabular} \\
\midrule
    4 & 
    \begin{tabular}[c]{@{}l@{}}
        $r_1$ \\ (Urea nitrogen, Urea nitrogen/creatinine)
    \end{tabular} &
    \begin{tabular}[c]{@{}l@{}}
        End-stage renal disease \\ Anemia \\ Chronic kidney disease \\
        Endometrial cancer \\ Myocardial infarction \\
        Ischemic heart disease \\ Arthritis \\ Diabetes \\
        Osteoporosis \\ Cataract \\ Glaucoma
    \end{tabular} &
    \begin{tabular}[c]{@{}l@{}}
        3.75 \\ 1.62 \\ 1.36 \\ 1.33 \\ 1.31 \\ 1.19 \\ 1.15 \\ 1.15 \\ 1.11 \\ 1.07 \\ 1.05
    \end{tabular} \\
\bottomrule
\multicolumn{4}{l}{
    \begin{tabular}[c]{@{}l@{}}
        \footnotesize $^{\dagger}$Dimension wherein the cluster exhibited the highest or lowest rates of aging \\
        \footnotesize $^\ddag$Diagnoses for which the cluster consistently achieved the highest odds ratios from first year to fourth year from the earliest date \\
        \footnotesize $^{\dagger\dagger}$Odds ratio calculated from data across four years
    \end{tabular}
}
\end{tabular}
\label{table:cluster_odds_future_diag}
\end{table}                           

\subsubsection{Health care expenditure of clusters}
\begin{figure}[ht]
	\centering
		\includegraphics[width=0.83\textwidth]{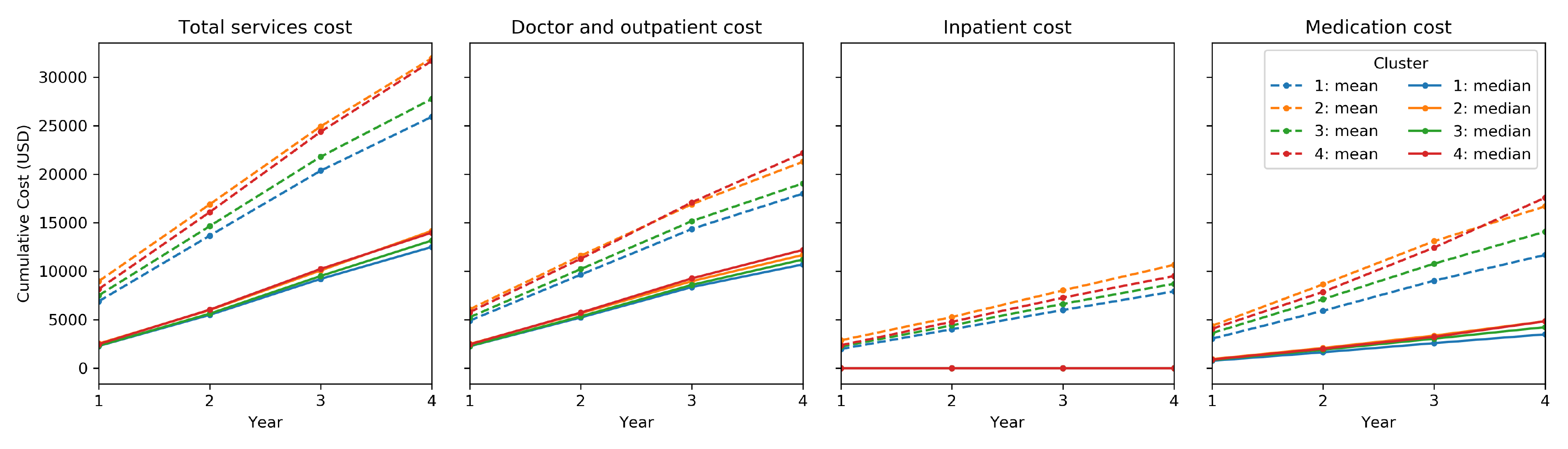}
	\caption{Mean and median of cumulative costs (US\$) by cluster and by cost type.}
	\label{FIG:cost_cluster}
\end{figure}

We saw that clusters 2 and 4 generally had higher spending based on mean and median overall health expenditures while cluster 1 had the lowest (Figure \ref{FIG:cost_cluster}). The mean cumulative costs of total services (doctor, outpatient, and inpatient) of an individual per cluster from first year to fourth year had the following values: for cluster 1, from US\$ 6,887.51 to US\$ 25,908.73; for cluster 2, from US\$ 8,954.16 to US\$ 31,934.47; for cluster 3, from US\$ 7,492.25 to US\$ 27,749.60; and for cluster 4, from US\$ 8,147.62 to US\$ 31,658.79. The results are likewise consistent when looking at costs for doctor and outpatient service types, as well as for inpatient admissions: in both services, it was clusters 2 and 4 that had the highest expenditures in general. Analysis of medication expenditures for the four clusters furthermore showed that the same two clusters had the greatest spending: the mean cumulative medication costs for clusters 2 and 4 across four years were US\$ 16,678.08 and US\$ 17,571.53, respectively. In contrast, for clusters 1 and 3, the average medication costs were US\$ 11,673.25 and US\$ 14,070.52. Conover's test of multiple comparisons confirmed that clusters 2 and 4 in general are significantly different when each cluster was compared to other clusters, but in most cases no significant difference was observed between the two.

\section{Discussion}
Our analyses show that the model was able to learn different dimensions of aging that are generally distinct and consistent with literature. The dimensions are highly correlated to lab components that are mostly unique from one another and these components are related to the historical diagnoses that are associated with their corresponding dimension. In the first dimension, the correlated components are BUN and BCR, and one of the associated diagnoses that is consistent in both historical and future events is diabetes. In the second dimension, TSH and CO$_2$ are the highly correlated components; heart failure, renal and pulmonary diseases, and hypothyroidism, are among the diagnoses associated with this dimension. In the third dimension, basophils are found to be related to hypothyroidism, asthma, anxiety, and heart diseases. In the fourth dimension, bilirubin is linked to alcohol abuse, atrial fibrillation, colorectal cancer, and Alzheimer's disease. 

We also described the different dimensions in terms of the aspects of aging they capture using findings we obtained from the analyses of correlations with lab components and associations to diseases. Dimension $r_1$ represents the `kidney rate of aging', given its correlation with kidney function components and kidney-regulated electrolytes as well as its association to kidney diseases. Dimension $r_2$ represents `thyroid rate of aging' because it is correlated with TSH and is associated to hypothyroidism and its comorbidities. Dimension $r_3$ represents `white blood cell rate of aging', since it correlates with leukocytes, particularly basophils. Finally, dimension $r_4$ represents `liver and heart rate of aging' because it is correlated with a liver function component while also being associated to liver and heart diseases. The difference in the aging dimensions extracted in this work and in the study by \citeauthor{pierson2019inferring} is primarily due to the types of features used. However, both used complete blood count and interestingly, both extracted a leukocyte subtype--basophils in our model and monocytes in theirs--as a strongly correlated component to an aging dimension.

The interpretability of the aging dimensions was further validated by the clustering analysis. We observed that the description of each cluster based on the related aging dimension--where the individuals belonging to the cluster showed the fastest rates of aging-agreed with the associated future diagnoses. Cluster 2, which comprises individuals aging the fastest along the thyroid dimension $(r_2)$, consistently has the highest odds for hypothyroidism and alcohol abuse among all clusters. Cluster 3, which is composed of individuals who progressed the fastest along the white blood cell (WBC) dimension $(r_3)$, has consistently achieved the highest odds of COPD and asthma. Cluster 4, which comprises individuals who exhibit the fastest rates of aging along the kidney dimension $(r_1)$, as expected, has the greatest odds of kidney diseases such as ESRD and CKD in addition to diabetes. 

\subsection{Relationship with health care expenditure}
Examination of healthcare costs revealed that fast kidney agers, fast thyroid agers, and fast liver and heart agers in $r_1$, $r_2$, and $r_4$ are statistically more likely to incur higher medical costs in the next four years compared to people with slower aging rates. This suggests that our model effectively represented the decline in these bodily functions, proxied by high future health care costs, through the aging rates it generated. Moreover, the kidney aging dimension has the strongest relationship with health care costs. Fast agers in $r_1$ are the most expensive among fast agers whether for services or medication. They are followed by fast agers in $r_2$ and $r_4$, who do not statistically differ in the bulk of incurred health care spending (doctor and outpatient costs, and in turn total services costs) but differ in smaller cost types, such that fast thyroid agers incur more medication costs while fast liver and heart agers incur more inpatient costs. The observation that high aging rates in the three dimensions relate to high health care utilization is supported by differences in spending among clusters, where we identified cluster 2 (highest $r_2$ and $r_4$ rates) and cluster 4 (highest $r_1$ rates) as the most expensive groups. This is expected since these clusters consistently had the highest odds for a wide range of chronic conditions and were also associated with chronic diseases with expensive treatments.  Particularly, we see in these clusters diseases that are known to have high direct health care costs, such as heart disease, cancer, stroke, Alzheimer's disease, diabetes, and CKD \citep{chronic-disease}. 

We note how results often support that the thyroid aging dimension and the liver and heart aging dimension are related. Many fast agers in $r_2$ are also fast agers in $r_4$, and these people are mainly grouped together in cluster 2. Even if we excluded the fast agers in both dimensions from the cost analysis, the bulk of costs for fast agers unique to $r_2$ and to $r_4$ are not statistically different. It is likely that our aging model somehow captured the complex relationship in health and disease between thyroid and liver, and as discussed previously, thyroid and heart. The liver has a crucial physiological role in thyroid hormone processes while thyroid hormones affect functional cell activities and metabolism in the liver \citep{Piantanida-2020}.

The white blood cell aging dimension has the weakest relationship with health care costs. Fast agers in $r_3$ are the least costly among fast agers in all cost types, and cluster 3 (with the highest $r_3$ rates) is among the least expensive among clusters. Costs of fast agers in $r_3$ are not so far apart from that of its slow agers, since more than half of these slow agers are nonetheless fast agers in other dimensions, potentially offsetting the anticipated savings in $r_3$ with medical costs not necessarily related to a decline in $r_3$. Particularly, many slow agers in $r_3$ are also fast agers in $r_4$, and vice versa. This is consistent with how $r_3$ is positively correlated to basophil count while $r_4$ is negatively correlated to this component, and how cluster 3 grouped together people with the highest $r_3$ rates and lowest $r_4$ rates.

Lastly, we note that individuals who are fast agers in simultaneous dimensions have higher expenditures than fast agers in only one dimension. This is illustrated by how the central tendencies of costs of fast agers for all dimensions decreased when the former group was removed. That high aging rates in multiple dimensions compound the necessary health care, whether for services or medication, further manifests how the latent dimensions generated by our model capture different aspects of aging. The information on how the aging dimensions and clusters relate to future costs could be useful in identifying individuals who must be given higher priority for preventive interventions--such as people whose associated aspect of aging would likely result to increased health care utilization. 

\subsection{Advantages of current work}
Modeling different dimensions of aging using routine laboratory tests allows future researchers to create larger cohorts without requiring specialized lab tests. Recall that our dataset is composed of 1.4 million individuals--five times larger than the population used by \citeauthor{pierson2019inferring} Our dataset is also more natural compared to the UK Biobank; individuals in our cohort did not undergo a large-scale recruitment process. Moreover, even though majority of the individuals with chronic conditions are over age 65, the age range of individuals in the Pierson dataset was limited to 40 to 69. Hence, by using a wider age range of 40 to 106 years our model captures more the apparent variability that exists among individuals as they age from midlife to late life. Lastly, we also expanded their definition of an individual's valid cross-sectional data since our work considers all the laboratory tests obtained within a month, which is a less restrictive criterion than all tests should be taken on a single day. Having these features in the dataset would allow future researchers to have more flexibility in reimplementing and deploying the model.

Besides the improvements made in the dataset, we modified the methodology of \citeauthor{pierson2019inferring} on how the model was optimized. Their methodology was extended by introducing additional metrics, aside from reconstruction correlation, that would make the model learn distinct aging dimensions while ensuring stability. With the use of intra-model similarity metric, the overlap in the description of the dimensions was minimized, leading to a model that maximized the different aspects of aging it had captured. Furthermore, our model generated even stronger correlations between the aging dimensions and components, with at least one feature per dimension having a coefficient beyond $\pm$0.60. Having said that, the refinements applied to both the dataset and the methodology have proven to be effective in producing interpretable aging dimensions that are in good agreement with literature.

\subsection{Limitations}
We note the following limitations in our work. Firstly, we only considered the data of Medicare and commercial members from a single large US healthcare insurer. We also prioritized practicality of the model for potential use in health care over comprehensiveness of aging dimensions to cover. The model and its possible applications are more accessible to people by using only features from inexpensive and nonspecialized tests. However, as a result we forewent capturing other common features of aging such as cognitive decline or pulmonary function. Furthermore, the process used to maximize the utility of the raw dataset led to the exclusion of other simple aging biomarkers such as weight and blood pressure in this study, which precluded a direct comparison of the derived dimensions with those identified by \citeauthor{pierson2019inferring} Further research is still needed to determine whether including such anthropometric measurements and vital signs, which are more monotonic with age, could significantly improve model performance.

Incorporating nuances of disease states (e.g., severity) is a critical factor to consider when designing interventions but is beyond the scope of this study. Moreover, interactions that may exist between diagnoses, laboratory tests, and/or aging dimensions were not captured. For instance, chronic diseases were treated as non-interacting variables in the OLS models when examining the association between the dimensions and the diagnoses. At the same time, biological processes are rarely independent and, as such, using our findings clinically should be deferred until further validation has been performed.

\section{Conclusion}
In this work, we demonstrated that the modifications we incorporated in the model of \citeauthor{pierson2019inferring} could be used to identify important factors of aging. We were able to validate their approach by implementing their model on a cohort of 1.4 million Americans. By using cross-sectional data from routine laboratory tests and by designing a framework of model selection which not only captures the quality of reconstruction but also the uniqueness of the dimensions represented by the latent variables, we uncovered four aging dimensions that are linked to 1) kidney, 2) thyroid, 3) white blood cells, and 4) liver and heart. We found associations supported by literature between these bodily functions and the chronic diseases that have the highest odds of developing among the fast agers in the corresponding aging dimension. We also discovered that fast agers in each dimension incur higher health care costs than slow agers, implying that the former will eventually require more health care in the future. The findings suggest that our model can effectively represent the decline in various bodily functions as rates of aging across different dimensions. 

The significance of our model lies in its practicality. It can determine multidimensional rates of aging without requiring longitudinal data and specialized tests. We can leverage this capability of our model to develop early intervention programs that can delay progression of chronic diseases, even for people lacking sufficient historical medical information. Additionally, the model has a potential application in designing prioritization strategies in health care due to the inherent association of rates of aging with both diseases and health care costs. 

\section{Data Availability}
The derived datasets generated for this study are available on request from OptumLabs at UnitedHealth Group, subject to restrictions imposed by relevant data protection laws such as the Health Insurance Portability and Accountability Act of 1996 (HIPAA). The raw de-identified laboratory and claims data from the UHG Clinical Discovery Database are proprietary and confidential information and are thus not publicly available.

\section{Appendix}
\subsection{Data Preprocessing}
The raw de-identified laboratory data of the preliminary cohort were cleaned prior deriving the dataset used for building the aging model. Standardization and regrouping were done on the 50 most common laboratory-tested human physical components, where standardization ensured that all results for a component are expressed in the same measurement unit while regrouping involved merging related components together or transferring subsets of values from one component to another. Cleaning also involved meticulously correcting encoding errors whenever there is sufficient evidence to assume irregularity and whenever batch correction is practical to do so. Afterwards, duplicates were removed as well as entries with zero values for components which could not possibly return a zero result. We then analyzed the distribution of results for each test and discarded outliers. To identify the outliers, we used an internal workflow which incorporated information on the critical values and healthy reference ranges of the components. In this workflow, we determined the upper (lower) bounds of the valid values by calculating the minimum (maximum) among the critical-value-based inter quartile range (IQR), the normal-range-based IQR, and the IQR derived from the whole data. 

\begin{figure}[ht]
	\centering
		\includegraphics[width=0.55\textwidth]{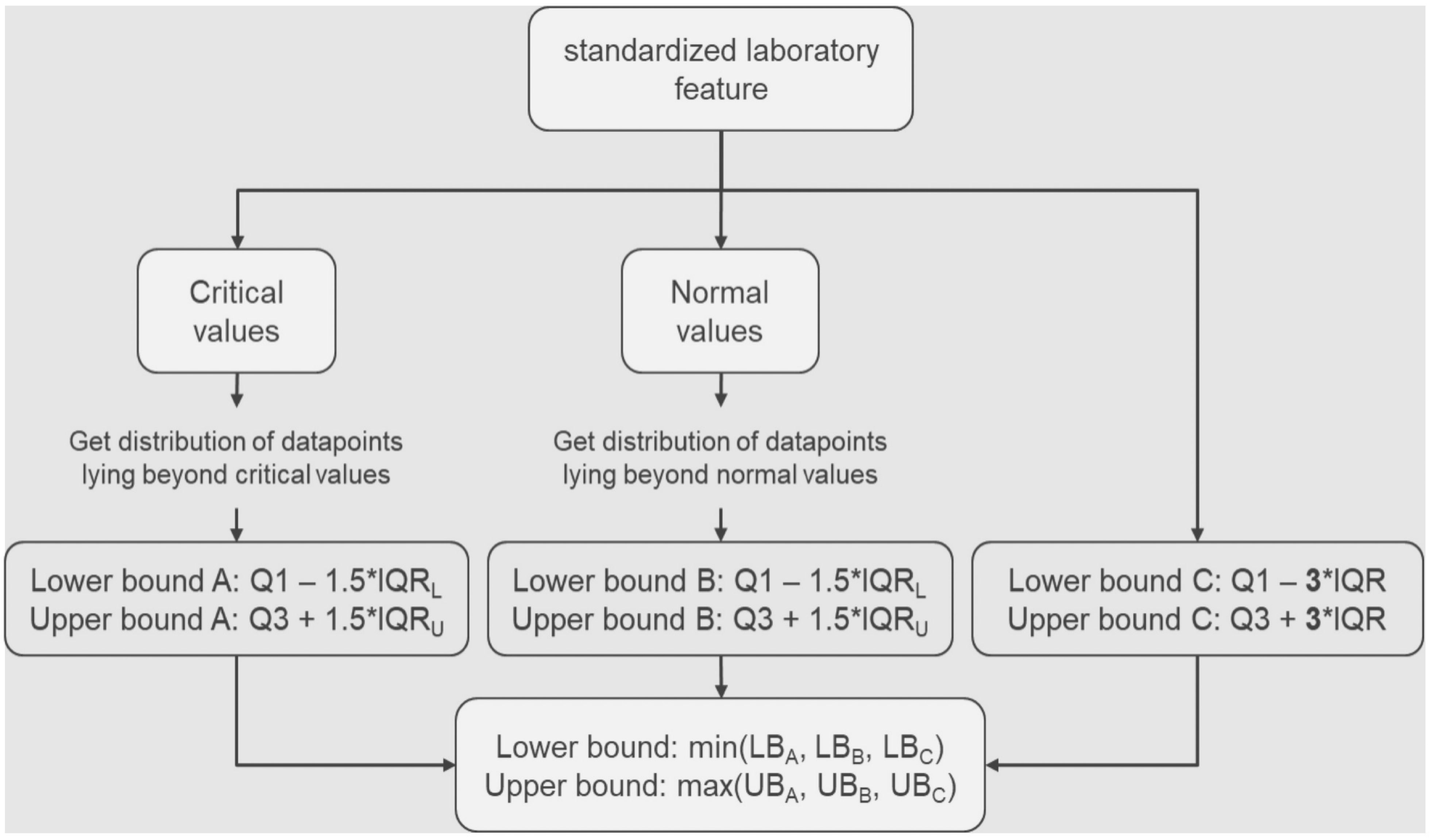}
	\caption{Flowchart describing the method we implemented to detect outliers. A laboratory result is removed if it went below the lower bound or above the upper bound.}
	\label{FIG:data_preprocessing}
\end{figure}

Critical-value-based IQR was calculated by first identifying the datapoints that were below or above the critical values. After identifying these datapoints, IQR was computed separately for the datapoints in lower and in upper regions. We denote the IQR for the lower region as IQRL while the IQR for the upper region as IQRU. From these, the lower and upper bounds LBA and UBA were calculated as in Figure \ref{FIG:data_preprocessing}. A similar procedure was done to calculate the bounds LBB and UBB from the normal-range-based IQR. In addition, we also computed for LBC and UBC incorporating a factor of $\pm$3.0 to the IQR for the entire distribution. After determining the final bounds, laboratory results in the lower (upper) region that were below (above) the identified cutoff are flagged as outliers. In the absence of critical or normal values, the algorithm defaulted to the 3.0 IQR rule. 

Only a few components in the database were tested from a substantial number of people. To build cross-sectional tables in this study without having to introduce uncertainty by imputing missing data, only individuals having complete results for the subset of components being considered may be included. With about 1.2 billion data entries left after outlier removal, we performed sensitivity analyses wherein we took note of how the cohort size diminishes with increasing number of components which are required to have at least one result on record – or complete cross-sectional data – on monthly, quarterly, and yearly aggregation periods. For multiple results of a person for the same component in the same aggregation period, the representative value was taken to be the latest, then highest, in the period. A monthly aggregation of 47 laboratory-tested components returned the best balance between time granularity and cohort size, as going beyond this number immediately shrinks the cohort size down to only about 160 thousand for monthly aggregation or about 300 thousand for yearly aggregation – which is just comparable to the cohort size of about 270 thousand used by \citeauthor{pierson2019inferring} The final derived dataset used for modeling in this study has 1,389,280 individuals equivalent to 65,296,160 test results across the 47 components shown in Table \ref{table:labfeatures}. For individuals having longitudinal data, we used the earliest available cross-sectional data so that the associations between model results and future diagnoses could be analyzed.

\begin{table}[ht]
    \scriptsize
    \caption{Laboratory-tested human components used as model features}
    \begin{tabular}{p{7cm}p{2.5cm}p{5cm}}
        \toprule
        Component & Standard Unit & Description \\
        \toprule
        Creatinine, ser/plas  & mg/dL & 
        \multirow{7}{5cm}{Kidney function tests; BUN and creatinine are part of basic metabolic panel} \\
        Urea Nitrogen (BUN), ser/plas & mg/dL & \\ 
        Urea Nitrogen/Creatinine (BCR), ser/plas & \% & \\
        Glomerular Filtration Rate (GFR; w/ factor for African American & mL/min per 1.73 m$^2$ & \\
        race, ser/plas/bld & body surface \\
        Glomerular Filtration Rate (GFR; w/o factor for African American & mL/min per 1.73 m$^2$ & \\
        race, ser/plas/bld & body surface \\
        \midrule
        Alanine Aminotransferase (ALT), ser/plas & U/L & \multirow{4}{5cm}{Liver function tests; part of comprehensive metabolic panel} \\
		Aspartate Aminotransferase (AST), ser/plas & U/L &  \\
		Alkaline Phosphatase (ALP), ser/plas & U/L & \\
		Bilirubin (total), ser/plas & mg/dL & \\
		\midrule
		Protein, ser/plas & g/dL & 
		\multirow{4}{5cm}{Protein tests; total protein and albumin are also part of liver function tests and comprehensive metabolic panel} \\ 
		Albumin, ser/plas & g/dL & \\
		Globulin, ser/plas & g/dL & \\
		Albumin/Globulin (A/G ratio), ser/plas & & \\
		\midrule
		Sodium, ser/plas & mEq/L & 
		\multirow{4}{5cm}{Blood electrolyte panel; part of basic metabolic panel} \\
		Chloride, ser/plas & mEq/L & \\
		Potassium, ser/plas & mEq/L & \\
		Carbon Dioxide, ser/plas & mEq/L & \\
		\midrule
		Calcium, ser/plas & mg/dL &
		\makecell[l]{Protein-attached calcium in blood; \\ part of basic metabolic panel} \\
		\midrule
		Glucose, ser/plas/bld & mg/dL  & Blood sugar; part of basic metabolic panel \\ 
		\midrule
		Platelets, bld & $10^3$ cells/$\mu$L &
		\makecell[l]{Total thrombocyte count per volume of \\ whole blood; part of complete blood count} \\
		\midrule
		Leukocytes, bld & $10^3$ cells/$\mu$L & 
		\makecell[l]{Total WBC count per volume of whole blood; \\ part of complete blood count} \\ 
		\midrule
		Basophils, bld  & $10^3$ cells/$\mu$L & 
		\multirow{5}{5cm}{WBC count (absolute units); part of complete blood count} \\
		Eosinophils, bld  & $10^3$ cells/$\mu$L &  \\
		Neutrophils, bld & $10^3$ cells/$\mu$L & \\
		Lymphocytes, bld & $10^3$ cells/$\mu$  & \\
		Monocytes, bld & $10^3$ cells/$\mu$L & \\
		\midrule
		Basophils/100 Leukocytes, bld & \% & 
		\multirow{5}{5cm}{WBC differential count (relative units)} \\
		Eosinophils/100 Leukocytes, bld & \% & \\
		Neutrophils/100 Leukocytes, bld & \% & \\
		Lymphocytes/100 Leukocytes, bld & \% & \\
		Monocytes/100 Leukocytes, bld & \% & \\ 
		\midrule
		Granulocytes immature, bld & $10^3$ cells/$\mu$L & \multirow{2}{5cm}{Underdeveloped basophils, eosinophils, and neutrophils in absolute or relative units} \\
		Granulocytes immature/100 Leukocytes, bld & \% & \\
		\midrule
		Erythrocytes, bld & $10^6$ cells/$\mu$L & 
		\makecell[l]{Total RBC count per volume of whole blood; \\ part of complete blood count} \\
		\midrule
		Hematocrit, bld & \% & 
		\makecell[l]{Packed cell volume; percentage of RBC in \\ whole blood} \\
		\midrule
		Hemoglobin, bld & g/dL & 
		\makecell[l]{Weight of oxygen-carrying protein molecule \\ per volume of RBC} \\
		\midrule
		Hemoglobin A1c/Hemoglobin (total), bld & \% & Percentage of glycated hemoglobin \\
		\midrule
		Erythrocyte Mean Corpuscular Volume (MCV), rbc & $\mu$m$^3$ & Average RBC size per cell \\
		\midrule
		Erythrocyte Mean Corpuscular Hemoglobin (MCH), rbc & pg & Average hemoglobin weight per RBC \\
		\midrule
		\makecell[l]{Erythrocyte Mean Corpuscular Hemoglobin \\ Concentration (MCHC), rbc} &
		g/dL & 
		\makecell[l]{Average hemoglobin concentration in a volume of \\ packed RBC; ratio of MCH to MCV} \\
		\midrule
		Erythrocyte Distribution Width, rbc & \% & 
		    \makecell[l]{Variability of RBC volume; percentage of one \\ standard deviation of RBC volume to MCV} \\
		\midrule
		Cholesterol (total), ser/plas & mg/dL &
		    \multirow{5}{5cm}{Lipid panel} \\
		Cholesterol in HDL, ser/plas & mg/dL & \\
		Cholesterol in LDL, ser/plas  & mg/dL & \\
		Cholesterol in VLDL, ser/plas & mg/dL & \\
		Triglyceride, ser/plas  & mg/dL & \\
		\midrule
		Thyrotropin, ser/plas & $\mu$IU/mL & Thyroid-stimulating hormone in blood \\
		\bottomrule
    \end{tabular}
	\label{table:labfeatures}
\end{table}

\clearpage

\subsection{Model Results}

\bibliographystyle{plainnat}

\begin{thebibliography}{16}
\providecommand{\natexlab}[1]{#1}
\providecommand{\url}[1]{\texttt{#1}}
\expandafter\ifx\csname urlstyle\endcsname\relax
  \providecommand{\doi}[1]{doi: #1}\else
  \providecommand{\doi}{doi: \begingroup \urlstyle{rm}\Url}\fi

\bibitem[Belsky et~al.(2020)Belsky, Caspi, Arseneault, Baccarelli, Corcoran,
  Gao, Hannon, Harrington, Rasmussen, Houts, and et~al.]{belsky_2020}
Daniel~W Belsky, Avshalom Caspi, Louise Arseneault, Andrea Baccarelli, David~L
  Corcoran, Xu~Gao, Eiliss Hannon, Hona~Lee Harrington, Line~Jh Rasmussen,
  Renate Houts, and et~al.
\newblock Quantification of the pace of biological aging in humans through a
  blood test, the dunedinpoam dna methylation algorithm.
\newblock \emph{eLife}, 9, 2020.
\newblock \doi{10.7554/elife.54870}.

\bibitem[Carmona and Michan(2016)]{carmona-2016}
Juan~Jos\'{e} Carmona and Shaday Michan.
\newblock Biology of healthy aging and longevity.
\newblock \emph{Revista de investigacion clinica; organo del Hospital de
  Enfermedades de la Nutricion}, 68, 2016.

\bibitem[{Centers for Medicare \& Medicaid Services}({n.d.})]{cmsccw}
{Centers for Medicare \& Medicaid Services}.
\newblock Chronic conditions data warehouse, {n.d.}
\newblock URL \url{https://www2.ccwdata.org}.

\bibitem[Denic et~al.(2016)Denic, Glassock, and Rule]{Denic-2016}
Aleksandar Denic, Richard~J. Glassock, and Andrew~D. Rule.
\newblock Structural and functional changes with the aging kidney.
\newblock \emph{Advances in Chronic Kidney Disease}, 23, 01 2016.
\newblock \doi{10.1053/j.ackd.2015.08.004}.

\bibitem[Ferrucci et~al.(2019)Ferrucci, Gonzalez-Freire, Fabbri, Simonsick,
  Tanaka, Moore, Salimi, Sierra, and de~Cabo]{Ferrucci-2019}
Luigi Ferrucci, Marta Gonzalez-Freire, Elisa Fabbri, Eleanor Simonsick, Toshiko
  Tanaka, Zenobia Moore, Shabnam Salimi, Felipe Sierra, and Rafael de~Cabo.
\newblock Measuring biological aging in humans: A quest.
\newblock \emph{Aging Cell}, 12 2019.
\newblock \doi{10.1111/acel.13080}.

\bibitem[Karol(2009)]{karol_2009}
Meryl~H. Karol.
\newblock How environmental agents influence the aging process.
\newblock \emph{Biomolecules and Therapeutics}, 17\penalty0 (2):\penalty0
  113–124, 2009.
\newblock \doi{10.4062/biomolther.2009.17.2.113}.

\bibitem[Kennedy et~al.(2014)Kennedy, Berger, Brunet, Campisi, Cuervo, Epel,
  Franceschi, Lithgow, Morimoto, Pessin, Rando, Richardson, Schadt, Wyss-Coray,
  and Sierra]{Kennedy-2014}
Brian~K. Kennedy, Shelley~L. Berger, Anne Brunet, Judith Campisi, Ana~Maria
  Cuervo, Elissa~S. Epel, Claudio Franceschi, Gordon~J. Lithgow, Richard~I.
  Morimoto, Jeffrey~E. Pessin, Thomas~A. Rando, Arlan Richardson, Eric~E.
  Schadt, Tony Wyss-Coray, and Felipe Sierra.
\newblock Geroscience: Linking aging to chronic disease.
\newblock \emph{Cell}, 159, 11 2014.
\newblock \doi{10.1016/j.cell.2014.10.039}.

\bibitem[Kim et~al.(2015)Kim, Kisseleva, and Brenner]{Kim-2015}
In~Hee Kim, Tatiana Kisseleva, and David~A. Brenner.
\newblock Aging and liver disease.
\newblock \emph{Current Opinion in Gastroenterology}, 31, 05 2015.
\newblock \doi{10.1097/mog.0000000000000176}.

\bibitem[Kirkwood(2005)]{Kirkwood-2005}
Thomas~B.L. Kirkwood.
\newblock Understanding the odd science of aging.
\newblock \emph{Cell}, 120, 2005.
\newblock \doi{10.1016/j.cell.2005.01.027}.

\bibitem[L\'{o}pez-Ot\'{i}n et~al.(2013)L\'{o}pez-Ot\'{i}n, Blasco, Partridge,
  Serrano, and Kroemer]{Lopez-Otin-2013}
Carlos L\'{o}pez-Ot\'{i}n, Maria~A. Blasco, Linda Partridge, Manuel Serrano,
  and Guido Kroemer.
\newblock The hallmarks of aging.
\newblock \emph{Cell}, 153, 06 2013.
\newblock \doi{10.1016/j.cell.2013.05.039}.

\bibitem[{National Center for Chronic Disease Prevention and Health
  Promotion}(2021)]{chronic-disease}
{National Center for Chronic Disease Prevention and Health Promotion}.
\newblock About chronic diseases, 2021.
\newblock URL \url{https://www.cdc.gov/chronicdisease/about/index.htm}.

\bibitem[Piantanida et~al.(2020)Piantanida, Ippolito, Gallo, Masiello, Premoli,
  Cusini, Rosetti, Sabatino, Segato, Trimarchi, Bartalena, and
  Tanda]{Piantanida-2020}
E.~Piantanida, S.~Ippolito, D.~Gallo, E.~Masiello, P.~Premoli, C.~Cusini,
  S.~Rosetti, J.~Sabatino, S.~Segato, F.~Trimarchi, L.~Bartalena, and M.~L.
  Tanda.
\newblock The interplay between thyroid and liver: implications for clinical
  practice.
\newblock \emph{Journal of Endocrinological Investigation}, 3 2020.
\newblock \doi{10.1007/s40618-020-01208-6}.

\bibitem[Pierson et~al.(2019)Pierson, Koh, Hashimoto, Koller, Leskovec,
  Eriksson, and Liang]{pierson2019inferring}
Emma Pierson, Pang~Wei Koh, Tatsunori Hashimoto, Daphne Koller, Jure Leskovec,
  Nicholas Eriksson, and Percy Liang.
\newblock Inferring multidimensional rates of aging from cross-sectional data,
  2019.

\bibitem[Sudlow et~al.(2015)Sudlow, Gallacher, Allen, Beral, Burton, Danesh,
  Downey, Elliott, Green, Landray, Liu, Matthews, Ong, Pell, Silman, Young,
  Sprosen, Peakman, and Collins]{Sudlow-2015}
Cathie Sudlow, John Gallacher, Naomi Allen, Valerie Beral, Paul Burton, John
  Danesh, Paul Downey, Paul Elliott, Jane Green, Martin Landray, Bette Liu,
  Paul Matthews, Giok Ong, Jill Pell, Alan Silman, Alan Young, Tim Sprosen, Tim
  Peakman, and Rory Collins.
\newblock Uk biobank: An open access resource for identifying the causes of a
  wide range of complex diseases of middle and old age.
\newblock \emph{PLoS Medicine}, 12, 3 2015.
\newblock \doi{10.1371/journal.pmed.1001779}.

\bibitem[Yoo et~al.(2017)Yoo, Kim, Cho, and Jee]{Yoo-2017}
Jinho Yoo, Yangseok Kim, Eo~Rin Cho, and Sun~Ha Jee.
\newblock Biological age as a useful index to predict seventeen-year survival
  and mortality in koreans.
\newblock \emph{BMC Geriatrics}, 17, 12 2017.
\newblock \doi{10.1186/s12877-016-0407-y}.

\bibitem[Zhang et~al.(2014)Zhang, Bai, Sun, Cai, Bai, Zhu, Zhang, and
  Chen]{W-2014}
W.~G. Zhang, X.~J. Bai, X.~F. Sun, G.~Y. Cai, X.~Y. Bai, S.~Y. Zhu, M.~Zhang,
  and Xiang-Mei Chen.
\newblock Construction of an integral formula of biological age for a healthy
  chinese population using principle component analysis.
\newblock \emph{The Journal of Nutrition, Health \& Aging}, 18, 02 2014.
\newblock \doi{10.1007/s12603-013-0345-8}.

\end{thebibliography}

\end{document}